\documentclass[letterpaper]{article} 
\usepackage{aaai2026}  
\usepackage{times}  
\usepackage{helvet}  
\usepackage{courier}  
\usepackage[hyphens]{url}  
\usepackage{graphicx} 
\usepackage{natbib}  
\usepackage{caption} 
\usepackage{algorithm}
\usepackage{algorithmic}

\usepackage{amsmath}
\usepackage{bm}
\usepackage{amssymb}
\usepackage{booktabs}
\usepackage{multirow}
\usepackage{pifont}
\usepackage{xcolor}
\usepackage[table]{xcolor}
\usepackage{pdfpages}

\usepackage{newfloat}
\usepackage{listings}
\DeclareCaptionStyle{ruled}{labelfont=normalfont,labelsep=colon,strut=off} 
\floatstyle{ruled}
\newfloat{listing}{tb}{lst}{}
\floatname{listing}{Listing}
\title{Sparse3DPR: Training-Free 3D Hierarchical Scene Parsing and Task-Adaptive Subgraph Reasoning from Sparse RGB Views}
\author {
    Haida Feng\textsuperscript{\rm 1,\rm 2}\equalcontrib,
    Hao Wei\textsuperscript{\rm 1}\equalcontrib,
    Zewen Xu\textsuperscript{\rm 1,\rm 2},
    Haolin Wang\textsuperscript{\rm 1,\rm 2},
    Chade Li\textsuperscript{\rm 1,\rm 2},
    Yihong Wu\textsuperscript{\rm 1,\rm 2}\thanks{Corresponding author.}
}
\affiliations {
    \textsuperscript{\rm 1}State Key Laboratory of Multimodal Artificial Intelligence Systems, \\ Institute of Automation, Chinese Academy of Sciences, Beijing, China. \\
    \textsuperscript{\rm 2}School of Artificial Intelligence, University of Chinese Academy of Sciences, Beijing, China \\
    fenghaida2024@ia.ac.cn, \{weihao2019@ia; xuzewen2020@ia; wanghaolin2023@ia; lichade2021@ia; yhwu@nlpr.ia\}.ac.cn
}

\usepackage{bibentry}

\begin{document}

\maketitle

\begin{abstract}
Recently, large language models (LLMs) have been explored widely for 3D scene understanding. Among them, training-free approaches are gaining attention for their flexibility and generalization over training-based methods. However, they typically struggle with accuracy and efficiency in practical deployment. To address the problems, we propose Sparse3DPR, a novel training-free framework for open-ended scene understanding, which leverages the reasoning capabilities of pre-trained LLMs and requires only sparse-view RGB inputs. Specifically, we introduce a hierarchical plane-enhanced scene graph that supports open vocabulary and adopts dominant planar structures as spatial anchors, which enables clearer reasoning chains and more reliable high-level inferences. Furthermore, we design a task-adaptive subgraph extraction method to filter query-irrelevant information dynamically, reducing contextual noise and improving 3D scene reasoning efficiency and accuracy. Experimental results demonstrate the superiority of Sparse3DPR, which achieves a 28.7\% EM@1 improvement and a 78.2\% speedup compared with ConceptGraphs on the Space3D-Bench. Moreover, Sparse3DPR obtains comparable performance to training-based methods on ScanQA, with additional real-world experiments confirming its robustness and generalization capability.
\end{abstract}


\section{Introduction}
Three-dimensional (3D) scene understanding is essential for embodied artificial intelligence, as it enables robots to understand, reason about, and execute natural language instructions within complex physical environments \cite{zhi2025lscenellm}. With the rapid advances of large language models (LLMs) \cite{liu2023summaryLLMs, achiam2023gpt4}, particularly their strong capabilities in communication, commonsense reasoning, and open-world knowledge integration, have motivated LLM-based solutions for 3D scene understanding. Existing approaches can be broadly classified into training-based and training-free methods. Training-based methods \cite{fu2024scene-llm, hong20233d, wang2023chat-3d} align 3D geometric or visual features with text features through specialized training, require complex architectures and incur high computational costs. Differently, training-free methods \cite{gu2024conceptgraphs, chandhok2024scenegpt} build explicit structured representations such as scene graphs (SGs) that encode objects and spatial relations, then convert SGs into textual context for the LLM, thereby leveraging its powerful zero-shot reasoning capability while eliminating training cost.

\begin{figure}[t]
    \centering
    \includegraphics[width=\columnwidth]{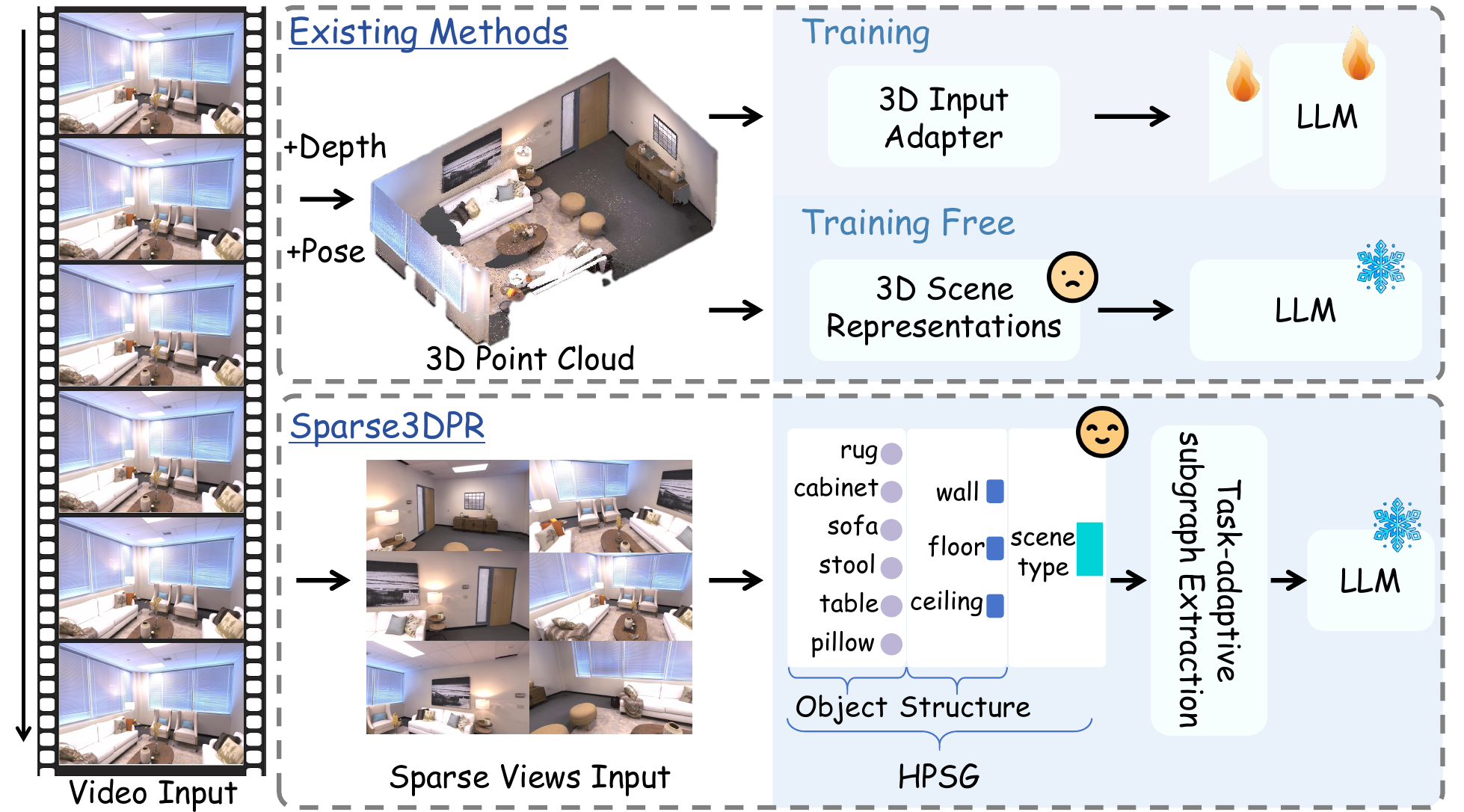}
    \caption{\textbf{Comparison of 3D scene understanding methods.} Unlike existing methods requiring dense 3D inputs or training, Sparse3DPR leverages sparse RGB views to construct a hierarchical plane-enhanced scene graph (HPSG) and performs task-adaptive subgraph extraction for efficient LLM‑based scene understanding.} 
    \label{fig:method-compare}
\end{figure}

Despite eliminating 3D-specific training, training-free methods still struggle with reasoning accuracy and computational efficiency in practical deployment. The first bottleneck is the quality and structural organization of 3D scene representations, which serve as the primary contextual input to LLMs and significantly impact reasoning reliability. Current SGs for LLM reasoning fall into two categories: flat SGs, such as ConceptGraphs \cite{gu2024conceptgraphs}, which encode all objects and pairwise relations in a single layer, lacking a hierarchy to organize their massive number of nodes, leading to redundant and token-inefficient inputs. Hierarchical SGs, like TB-HSU \cite{xu2025tbHSU}, group objects by their functionality. While offering more structure, their hierarchy disrupts the scene's natural spatial proximity, breaking physical coherence and introducing reasoning ambiguity. Another critical issue is the contextualization method for reasoning. The approach in prior work, such as SceneGPT \cite{chandhok2024scenegpt}, adopts a static, non-adaptive approach that converts the entire unfiltered scene representation into a single context for each query. This approach increases computational cost and degrades reasoning accuracy, as task-relevant details are easily obscured by irrelevant information.

To address the challenges, we propose Sparse3DPR, a novel training-free framework for 3D scene understanding from sparse-view RGB inputs, as illustrated in Figure~\ref{fig:method-compare}. Inspired by human cognitive mechanisms for organizing scenes \cite{epstein1998cortical}, Sparse3DPR constructs a hierarchical plane-enhanced scene graph (HPSG) with a geometrically grounded hierarchy anchored by dominant planes (e.g., walls, floors, ceilings), achieving structural efficiency and spatial coherence to overcome the limitations of prior SGs. Semantically, it leverages pretrained vision-language models (VLMs) to enrich fine-grained objects with rich concepts, thereby enabling open-vocabulary capabilities. Furthermore, to effectively leverage this rich and effective representation for LLM-based reasoning, we introduce a task-adaptive subgraph extraction method. It mimics human selective attention by dynamically retrieving only the query-relevant parts of the HPSG. This reduces contextual noise and enables more accurate and efficient task-specific reasoning. To validate the effectiveness of Sparse3DPR, we evaluate it on 3D question answering (QA), a representative task that assesses its unified ability to comprehend language, perceive semantics, and reason about complex spatial relationships within 3D scenes. We further report the average inference time per query as a metric of reasoning efficiency.

The primary contributions of Sparse3DPR are succinctly summarized as follows:
\begin{itemize}
    \item We design a hierarchical plane-enhanced scene graph that incorporates dominant planar structures as the spatial anchors, which provides a more human-intuitive representation, achieves 6.7\%/13.8\% EM@1 improvement and 13.0\%/7.8\% faster than flat scene graphs~\cite{gu2024conceptgraphs} and affordance-based hierarchical scene graphs~\cite{xu2025tbHSU} on the 3D QA task.
    
    \item We propose a novel task-adaptive subgraph extraction method that dynamically prunes the full scene graph to generate a query-relevant subgraph, which provides the LLM with a focused, noise-free context, thereby enhancing 3D scene reasoning speed and accuracy.
    
    \item By integrating the above innovations, we present a training-free scene understanding framework that requires sparse-view RGB images solely. Our method achieves an improvement of 28.7\% EM@1 and a speedup of 78.2\% compared to ConceptGraphs, while demonstrating competitive performance on public benchmarks and generalizability in real-world scenes.

\end{itemize}

\section{Related Work}
\subsection{3D Scene Representations for LLM Reasoning}
Recent works using dense 3D representations (e.g., CLIP-based \cite{radford2021CLIP,peng2023openscene, mohiuddin2024opensu3d, jatavallabhula2023conceptfusion, yamazaki2024openfusion}) support open-vocabulary perception but lack structure, hindering compositional reasoning. In contrast, 3D SGs offer a structured, object-centric alternative. Classic methods \cite{armeni20193dscenegraph, wald2020learning3dgraph} build graphs where nodes represent objects or spatial regions and edges encode spatial or semantic relationships. Hydra \cite{hughes2022hydra, hughes2024hydra2} extends this to real-time hierarchical SGs that incrementally organize environments from high-level structures (e.g., buildings and rooms) to fine-grained entities (e.g., objects and agents). These representations effectively organize spatial and are well-suited for LLM integration, thanks to their structured format and ease of conversion to natural language. For example, TB-HSU \cite{xu2025tbHSU} and ConceptGraphs \cite{gu2024conceptgraphs} construct SGs that can be integrated with LLMs for reasoning. However, TB-HSU is constrained to a predefined set of semantic categories, which limits its generalization and reasoning flexibility. While ConceptGraphs is capable of building open-vocabulary SGs, its flat graph structure lacks a meaningful hierarchy, limiting efficient and coherent reasoning in LLMs. In contrast, our proposed HPSG incorporates dominant planar structures (walls, floors, ceilings) as spatial anchors, while supporting open-vocabulary capabilities and forming a spatially coherent and semantically rich hierarchy suited for LLM reasoning.

\subsection{3D Scene Understanding with LLMs}
Recent works integrate LLMs into 3D scene understanding, often by training them to align spatial and textual representations on 3D-specific datasets~\cite{hong20233d, wang2023chat-3d, huang2024chatscene, chen2024ll3da, zhang2025flatland, fu2024scene-llm, chen2024grounded-3d-llm, spatiallm}. For instance, Scene-LLM~\cite{fu2024scene-llm} jointly trains 3D encoders with LLMs, while LL3DA~\cite{chen2024ll3da} uses fusion modules to align geometric and textual features. Despite their effectiveness, these methods require large-scale 3D datasets and still struggle to fully align geometric structures with natural language representations. Alternatively, training-free approaches~\cite{mohiuddin2024opensu3d, gu2024conceptgraphs, chandhok2024scenegpt} leverage pretrained LLMs and scene representations, enabling reasoning without the need for 3D-specific training. SceneGPT~\cite{chandhok2024scenegpt}, for instance, employs SGs built from dense RGB-D inputs for LLM-based scene understanding. However, converting the entire SG into a single prompt risks exceeding the LLM's context window, and it also interferes with its reasoning by introducing task-irrelevant information. In contrast, our framework is training-free and relies only on sparse-view RGB images, which eliminates the need for 3D-specific data. Furthermore, its task-adaptive subgraph extraction method provides the LLM with more compact and relevant information by pruning away task-irrelevant entities from the full SG.

\section{Methodology}
\begin{figure*}[!t]
    \centering
    \includegraphics[width=\linewidth]{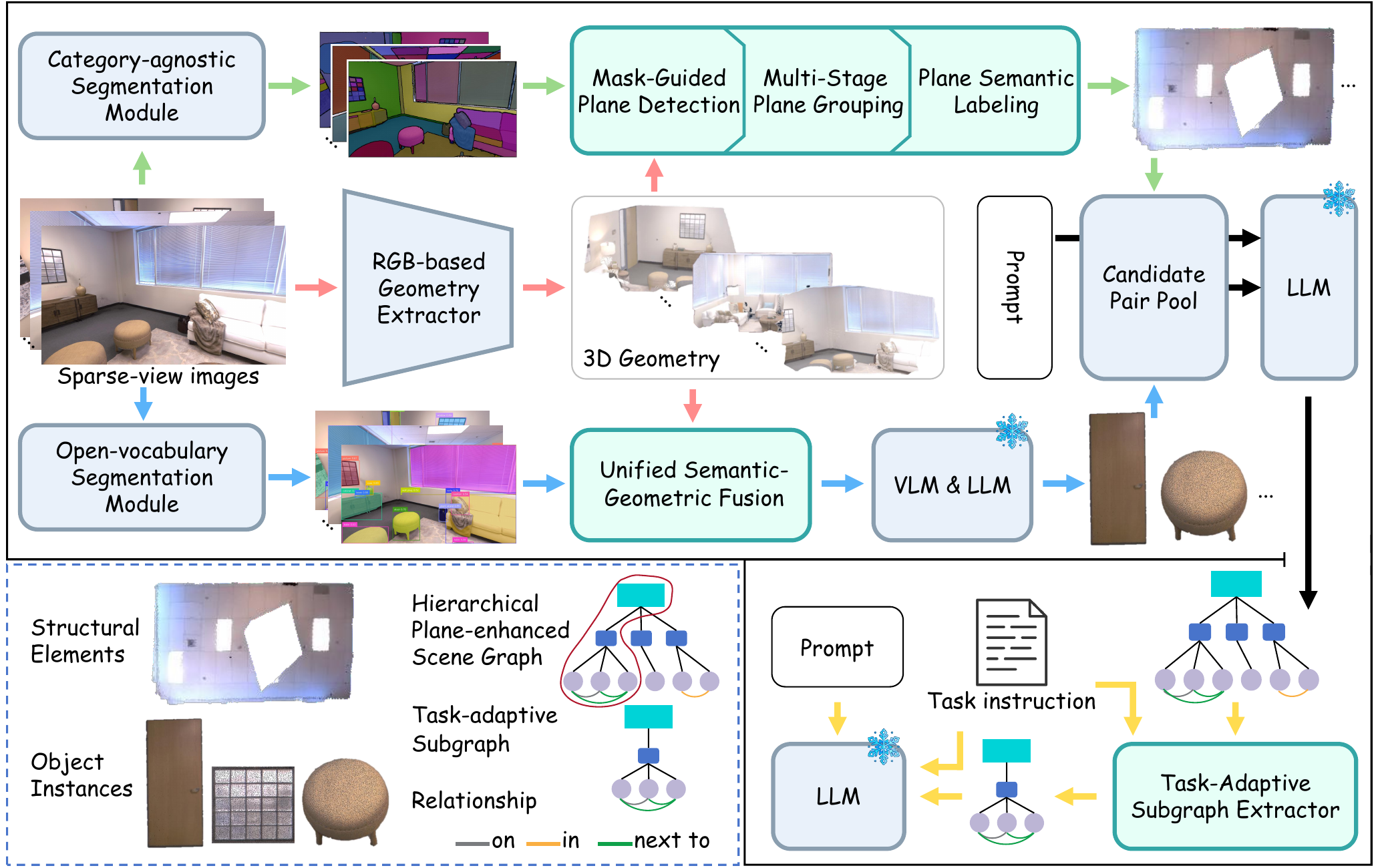}
     \caption{\textbf{Overview of Sparse3DPR framework.} Sparse3DPR is a training-free framework for 3D scene understanding from sparse RGB views. It parses the scene through two branches that integrate 3D geometry: one extracts structural elements by applying class-agnostic masks combined with structural plane detection and labeling, while the other uses open-vocabulary masks for semantic-geometric fusion to obtain object instances and generate captions. These components form a candidate pair pool of topological connections, which are further refined by an LLM to estimate spatial relations between object pairs and construct the HPSG. A task-adaptive subgraph extractor then selects relevant context from the HPSG for LLM-based reasoning.}
    \label{fig:method}
\end{figure*}
The proposed Sparse3DPR framework provides a training-free solution for 3D scene understanding from sparse RGB views. It first parses the scene to construct an HPSG as a compact and reasoning-oriented scene representation, and then dynamically prunes the HPSG to focus on query-relevant context via task-adaptive subgraph extraction. The overview of the framework is shown in Figure \ref{fig:method}.

\subsection{Hierarchical Scene Parsing}
\subsubsection{Geometry Extraction from Sparse RGB Inputs.}
Effective hierarchical scene parsing requires consistent and geometrically reliable information. Instead of requiring dense, pose-aligned RGB-D inputs, Sparse3DPR utilizes DUSt3R \cite{wang2024dust3r}, a learning-based multi-view reconstruction framework that can infer scene geometry directly from sparse monocular RGB images, even under challenging wide-baseline conditions. To ensure sufficient spatial coverage and viewpoint diversity, we uniformly sample a sparse subset of input images from the available pool, denoted as $\mathcal{I} = \{I_i\}_{i=1}^{n}$. The geometry inference is then formulated as a parameterized function $\Phi_{\text{D3R}}$:
\begin{equation}
(\mathcal{X}, \mathcal{C}) = \Phi_{\text{D3R}}(\mathcal{I}; \boldsymbol{\theta}_{\text{D3R}}), \quad P_i = X_i \odot C_i,
\end{equation}
where $\mathcal{X} = \{X_i\}_{i=1}^{n}$ denotes the set of per-view 3D point maps and $\mathcal{C} = \{C_i\}_{i=1}^{n}$ represents the corresponding confidence maps. $\boldsymbol{\theta}_{\text{D3R}}$ denotes the tunable parameters of $\Phi_{\text{D3R}}$. Each $X_i \in \mathbb{R}^{W\times H\times 3}$ encodes the 3D coordinates of every pixel, while each $C_i \in \mathbb{R}^{W\times H}$ provides the associated per-pixel confidence scores. The element-wise product $\odot$ is used to suppress low-confidence points, resulting in a filtered point cloud $P_i \in \mathbb{R}^{N_i \times 3}$ for each view. 

\subsubsection{Structural Element Extraction.} This section details our three-stage pipeline for robustly extracting the foundational structure of the scene (walls, floors, ceilings) from the noisy geometry recovered from sparse views. The pipeline progressively refines an initial set of plane hypotheses into a globally consistent set of structural elements.

The process begins with mask-guided plane detection, where we first leverage SAM2~\cite{ravi2024sam2} to generate a set of class-agnostic segmentation masks $\mathcal{M}_i = \{M_i^j\}_{j=1}^{m_i}$ for each input image $I_i$, with $m_i$ being the number of masks. Each 2D mask $M_i^j$ is lifted into 3D as a candidate point cloud $P_i^j$ by applying it to the inferred point map $X_i$, weighted by the per-pixel confidence map $C_i$:
\begin{equation}
P_i^j = X_i \odot (M_i^j \odot C_i).
\end{equation}
We then by applying a RANSAC-based plane fitting algorithm $\mathcal{R}_{plane}$ to each candidate point cloud $P_i^j$ to generate an initial set of candidate planes:
\begin{equation}
    (\mathbf{n}_i^j, d_i^j, \mathcal{O}_i^j) = \mathcal{R}_{plane}(P_i^j; \tau_{\text{dist}}, \rho_{\text{min\_inlier}}),
\end{equation}
where $\mathbf{n}_i^j$ and $d_i^j$ are the unit normal and offset of the estimated plane, and $\tau_{\text{dist}}$ is the inlier distance threshold. A candidate plane $\Pi_i^j$ is considered valid if its inlier ratio $|\mathcal{O}_i^j|/|P_i^j|$ exceeds $\rho_{\text{min\_inlier}}$, where the corresponding inlier set $\mathcal{O}_i^j$ is given by:
\begin{equation}
    \mathcal{O}_i^j = \{ \mathbf{p} \in P_i^j \mid |\langle \mathbf{n}_i^j, \mathbf{p} \rangle - d_i^j| \leq \tau_{\text{dist}}\}.
\end{equation}

To merge the initially fragmented and inconsistent candidate planes into a globally coherent set, our pipeline next performs multi-stage plane grouping. This begins with intra-view refinement for each view independently, where we first cluster local candidate planes using DBSCAN in the plane parameter space (PPS), which is defined by the estimated plane parameters $(\mathbf{n},d)$. This step yields coarse groupings of geometrically co-planar surfaces. To enhance spatial continuity and completeness, these coarse groups are then expanded by a geometry-aware region growing module that incorporates nearby points satisfying the following strict angular and distance constraints: 
\begin{equation}
\begin{aligned}
\cos^{-1}(\mathbf{n}_i^j{}^\top \mathbf{n}_{\mathbf{p}}) < \theta_{\text{ang}}, \
|\mathbf{n}_i^j{}^\top \mathbf{p} - d_i^j| < \delta_{\text{dist}},
\end{aligned}
\label{eq:region_growing}
\end{equation}
where $\mathbf{n}_{\mathbf{p}}$ is the pre-computed normal of point $\mathbf{p}$, and $\theta_{\text{ang}}$ and $\delta_{\text{dist}}$ are the respective angular and distance thresholds. While this expansion effectively recovers fragmented regions, it can introduce residual over-segmentation. Therefore, a final intra-view DBSCAN is applied to consolidate the expanded planes and merge adjacent, co-planar segments, yielding a refined set of planes $\Pi_i^{\prime}$ for each view. These refined per-view sets are then fused in the cross-view alignment stage, where we aggregate all refined sets $\{\Pi_i^{\prime}\}$ and perform a global DBSCAN clustering in the PPS. This process robustly associates and merges observations of the same plane across different views to produce the globally consistent set of distinct planar surfaces $\Pi_{\text{global}}$.

The final stage of our pipeline is plane semantic labeling. To distinguish structural elements within $\Pi_{\text{global}}$, we introduce a geometry-driven semantic labeling method that assigns each plane to a structural category, such as floor, wall, or ceiling, or as a non-structural surface. We begin by detecting floor planes, which are identified by their normal vectors that are closely aligned with the gravity direction, thereby establishing a gravity-aligned coordinate frame for the scene. Using this frame, ceiling planes are classified based on their normals forming an angle of less than $20^{\circ}$ with the gravity direction and exhibiting a positive offset $d$. Wall candidates are identified by normals approximately orthogonal to the floor plane. To distinguish true structural walls from other large vertical surfaces (e.g., cabinet sides), these candidates are then validated through a multi-criteria filtering process. This filter assesses geometric attributes, such as planar area and boundary length, and requires at least two supporting observations from different viewpoints to ensure geometric consistency.

\subsubsection{Object Instance Extraction.}
We introduce an object-centric pipeline that robustly extracts globally consistent 3D object instances from sparse RGB views. It further enriches each instance with descriptive captions and semantic tags.

The process begins with an open-vocabulary instance segmentation module. For each sparse-view image $I_i$, we first apply the RAM++ model $\mathcal{F}_{\text{T}}$~\cite{huang2023ram++} to predict a set of open-vocabulary category labels. These labels are used as text prompts for the GroundingDINO detector $\mathcal{F}_{\text{D}}$~\cite{liu2024groundingdino} to localize candidate object regions. The detected regions are then passed to the SAM2 segmenter $\mathcal{F}_{\text{S}}$ to generate instance masks. To ensure cross-view consistency, we introduce a propagation module $\mathcal{F}_{\text{P}}$~\cite{cheng2023DEVA}, which associates masks of the same object across multiple views. Through this segmentation process, the module generates cross-view consistent instance masks, where each mask is assigned a persistent ID to maintain identity consistency for the object across different viewpoints. The mask generation and ID assignment are jointly formulated as:
\begin{equation}
\begin{aligned}
\big\{ (M_i^j, L^j) \big\}_{j=1}^{m_i} = \mathcal{F}_{\text{P}} \left( \mathcal{F}_{\text{S}} \left( \mathcal{F}_{\text{D}}(I_i, \mathcal{F}_{\text{T}}(I_i)) \right), \mathbf{S}_{i-1} \right),
\end{aligned}
\end{equation}
where $m_i$ is the number of detected instances in view $i$, $M_i^j$ is the mask for the $j$‑th instance, and $L^j$ is the consistent instance ID shared across views. $\mathbf{S}_{i-1}$ is the propagated state from previous views used to ensure ID consistency.

Building upon the outputs of the previous stage, we perform a unified semantic-geometric fusion. The process begins by combining the 2D object masks from view $i$ with their corresponding 3D geometry from DUSt3R, which yields a set of local object candidates $\hat{o}_i=\{\hat{o}_i^{(1)},\ldots,\hat{o}_i^{(m)}\}$, where $m$ denotes the number of detected objects. Each candidate $\hat{o}_i^{(m)}=\langle P_i^{m},L^{m}\rangle$ comprises its 3D geometry $P_i^{m}$ and an associated instance ID $L^{m}$. To suppress outliers caused by imprecise mask boundaries, these candidates are subsequently filtered using a density-based clustering method (e.g., DBSCAN), retaining only the dominant cluster as the final geometry. The refined candidates are then progressively merged into a global object set $O = \{ o^{(1)}, \ldots, o^{(K)} \}$ based on a unified semantic-geometric association rule. A new detection is merged with an existing object if their IDs match, or if their 3D IoU $\varphi_{\text{geo}}$ exceeds a threshold $\kappa$. Otherwise, the detection is registered as a new instance. This fusion logic is formally expressed as:
\begin{equation}
O \leftarrow
\begin{cases}
\left( O \setminus \{ o^{(k)} \} \right) \cup 
\left\{ \langle P^{k} \cup P_{i}^{m}, L^{k} \rangle \right\} \\
\quad \text{if } \exists\, k ,\ \varphi_{\text{geo}} > \kappa \lor L^{m} = L^{k}, \\
O \cup \left\{ \langle P_{i}^{m}, L^{m} \rangle \right\} \quad \text{otherwise.}
\end{cases}
\end{equation}

With the global 3D object set established, the final stage performs vision-language caption generation. We employ a two-stage procedure that first uses a VLM to generate preliminary captions, which are then refined by an LLM. For each object, the process begins by selecting the top-5 views with the highest segmentation confidence. The corresponding object-centered image crops are fed to the VLM with the prompt “\textit{Provide a concise description of the main object in this image.}”. This yields a set of preliminary view-specific captions $\widetilde{c}_v = \{ \widetilde{c}_{v,1}, \widetilde{c}_{v,2}, \ldots, \widetilde{c}_{v,5} \}$. These preliminary captions are then refined and consolidated by the LLM using a prompt template $\mathcal{P}$, producing a coherent final caption $c_v$ together with a canonical tag $t^*$ and a candidate tag set $\mathcal{T}$:
\begin{equation}
    (c_v, t^*, \mathcal{T}) = \operatorname{LLM}(\widetilde{c}_{v}, \mathcal{P}).
\end{equation}

\subsubsection{HPSG Construction.} Given the set of scene components, including structural plane elements $\Pi$ and object instances $O$, we compute a similarity matrix $\mathbf{S}$, where each entry is defined as the 3D bounding box IoU between every pair of components. A minimum spanning tree (MST) is then estimated from $\mathbf{S}$ to establish a candidate pool of topological connections among all components. We further estimate the spatial relations such as 'on', 'in', and 'next\_to' between object instance pairs found within the candidate pool by feeding their captions and 3D positions to the LLM using a prompt template. In addition, the LLM summarizes captions of all $O$ to infer the global scene type (e.g., office or room), which is then combined with each structural label to generate captions for $\Pi$, such as “\textit{This is a \{wall\} in the \{office\}.}” We then construct the HPSG, denoted as $\mathcal{G} = (\mathcal{V}, \mathcal{E})$. The node set $\mathcal{V} = \bigcup_{l=0}^{2} \mathcal{V}_l$ comprises three levels of scene components: $\mathcal{V}_0$ corresponds to the global scene type (e.g., office), $\mathcal{V}_1$ to structural plane elements, and $\mathcal{V}_2$ to object instances. The edge set $\mathcal{E} = \bigcup_{m=0}^{2} \mathcal{E}_m$ captures the hierarchical and spatial relationships, including scene type-to-structure default links $\mathcal{E}_0$, structure-to-object topological connections $\mathcal{E}_1$, and spatial relations between objects $\mathcal{E}_2$.

\subsection{Task-Adaptive Subgraph Extraction}
To dynamically extract task-relevant subgraphs from the full HPSG, we use each node’s caption as a semantic anchor for identifying seed nodes. All node captions $\{e_i\}_{i=1}^{N}$ and the user query $q$ are encoded into their corresponding embeddings using a pre-trained SentenceTransformer $S(\cdot)$ model \cite{reimers2019sentencebert}:
\begin{equation}
\mathbf{F}_{e_i} = S(e_i), \quad \mathbf{F}_q = S(q). 
\end{equation}
The semantic alignment between the query and each node is then measured using a query-aware scoring function:
\begin{equation}
S(q, e_i; \tau) = \exp \left( \frac{\mathcal{N}(\mathbf{F}_q) \cdot \mathcal{N}(\mathbf{F}_{e_i})}{\tau} \right),
\end{equation}
where $\mathcal{N}$ denotes L2 normalization, and $\tau$ is a temperature scalar (set to 0.07). To efficiently select the most relevant nodes, we use FAISS \cite{douze2024faiss} to retrieve the top-$K$ seed nodes that maximize the alignment score:
\begin{equation}
I^* = \operatorname*{arg\,topK}_{i \in \{1, \ldots, N\}} (\mathcal{S}(q, e_i;\tau)).
\end{equation}
To retain sufficient contextual information while preserving the hierarchical structure, the retrieved seed nodes are further expanded with their directly connected neighbors $\mathcal{N}(I^*)$ and second-order neighbors $\mathcal{N}^2(I^*)$, together with their corresponding edges from the original graph. The resulting localized subgraph is defined as:
\begin{equation}
\mathcal{G}^*_q = \mathcal{G}\big[ I^* \cup \mathcal{N}(I^*) \cup \mathcal{N}^2(I^*) \big],
\end{equation}
which serves as a compact task-focused subgraph that preserves essential contextual information.

\section{Experiments}
\subsection{Main Results}
We design a multi-perspective evaluation to validate the effectiveness of our framework. We first quantitatively evaluate the accuracy and consistency of the nodes in our HPSG via a 3D semantic understanding experiment, as these nodes serve as the contextual foundation for downstream tasks. We then evaluate the framework's reasoning and understanding capabilities using the 3D QA task, which evaluates its ability to perform spatial and semantic reasoning grounded in complex 3D environments via natural language question answering. Finally, we conduct qualitative experiments in a real-world lab scene to further validate the generalization and practical applicability of Sparse3DPR.

\begin{table}[ht]
    \centering
    \large
    \renewcommand{\arraystretch}{1.1}  
    \resizebox{\columnwidth}{!}{%
    \begin{tabular}{l|l|cc}
        \toprule
        \textbf{Supervision} & \textbf{Method} & \textbf{F-mIoU (\%)} $\uparrow$ & \textbf{mAcc (\%)} $\uparrow$ \\
        \midrule
        \multirow{4}{*}{\textbf{Training}}
        & CLIP (rd64-uni) & 39.84 & 28.21 \\
        & CLIP (rd64-uni-refined) & 13.00 & 13.00 \\
        & LSeg & 51.54 & 33.39 \\
        & OpenSeg & 53.74 & 41.19 \\
        \midrule
        \multirow{8}{*}{\textbf{Training Free}}
        & MaskCLIP & 0.94 & 4.53 \\
        & Mask2Former + Global CLIP feat  & 13.11 & 10.42 \\
        & ConceptFusion & 31.31 & 24.16 \\
        & ConceptFusion + SAM & 38.70 & 31.53 \\
        & HOV-SG + ViT-H-14 & 38.60 & 30.40 \\
        & ConceptGraphs$^\text{R}$ & 34.68 & 37.52 \\
        & ConceptGraphs-Detector$^\text{R}$ & 34.70 & \textbf{37.97} \\
        & \cellcolor[gray]{0.95}\textbf{Sparse3DPR (Ours)} 
        & \cellcolor[gray]{0.95}\textbf{39.71} 
        & \cellcolor[gray]{0.95}35.12 \\
        \bottomrule
    \end{tabular}
    }
    \caption{\textbf{3D semantic understanding on Replica.} Results marked with $^\text{R}$ indicate those reproduced by our implementation. All other results are quoted from their respective publications. Training-based methods involve task-specific fine-tuning, while training-free methods leverage pre-trained models without this requirement.}
    \label{fig:semantic_seg_table}
\end{table}

\subsubsection{3D Semantic Understanding.} We evaluate the 3D semantic understanding on the Replica \cite{straub2019replica}, which contains five office scenes and three apartment rooms. As shown in Table~\ref{fig:semantic_seg_table}, Sparse3DPR is compared with several training-based methods \cite{luddecke2022CLIPSeg, li2022Lseg, ghiasi2022openseg} and training-free methods \cite{zhou2022maskclip, jatavallabhula2023conceptfusion, werby2024HOV-SG, gu2024conceptgraphs}. Sparse3DPR achieves the highest F-mIoU among all training-free baselines and even outperforms several training-based approaches, while also delivering competitive mAcc scores. This performance stems from the semantic-geometric fusion employed during hierarchical scene parsing. Sparse3DPR first establishes a robust geometric foundation by applying DBSCAN to suppress noise in the initial segments, then progressively aligns cross-view features under both geometric and semantic constraints. This process yields a globally consistent scene representation which is essential for supporting precise and reliable reasoning in downstream 3D scene understanding tasks.

\begin{figure*}[ht]
    \centering
    \includegraphics[width=\linewidth]{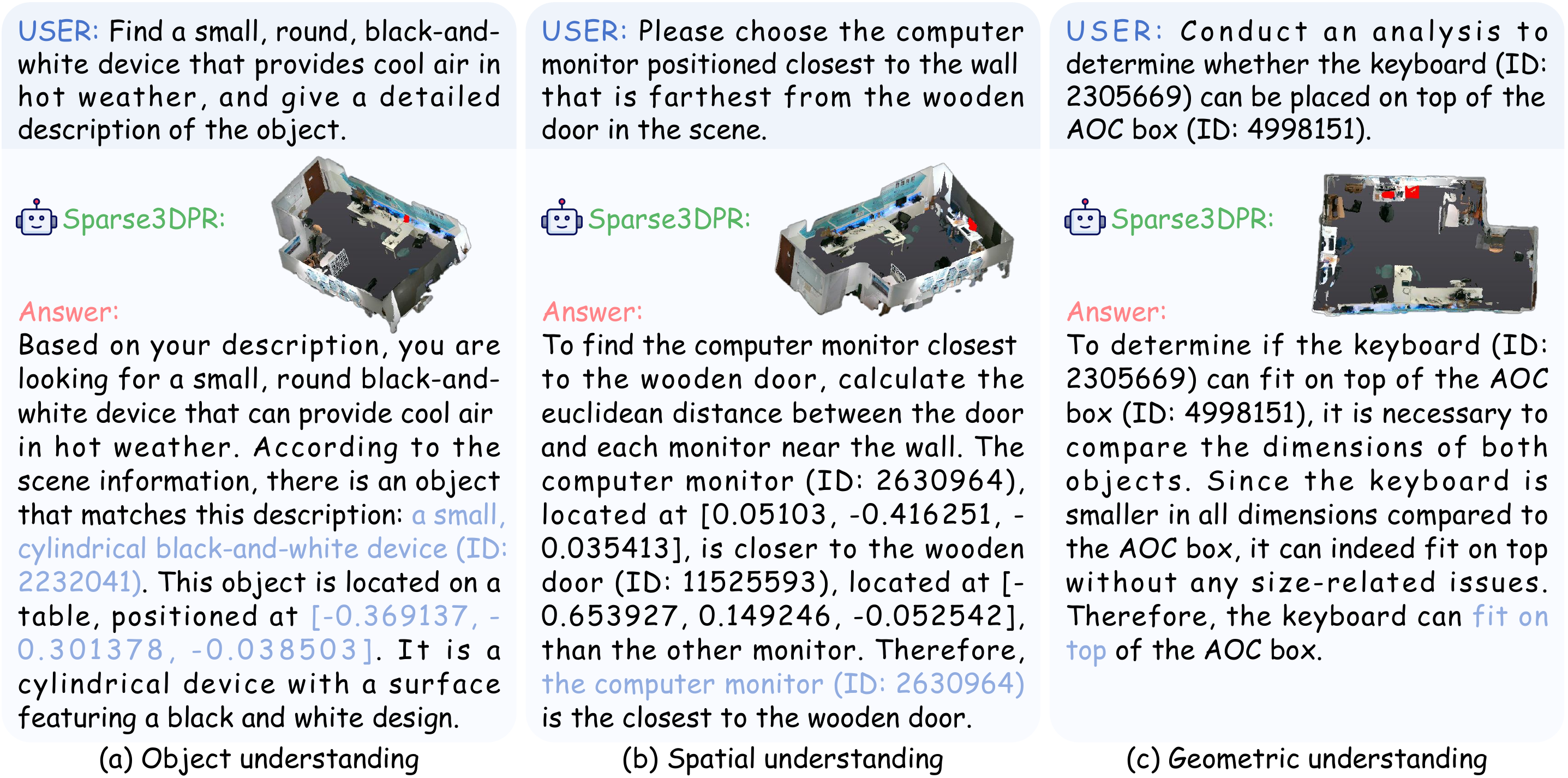} 
    \caption{\textbf{Qualitative results.} We showcase Sparse3DPR performing object, spatial, and geometric understanding in a lab scene and locating target objects within the scene. These examples demonstrate its ability to adapt to diverse tasks and generalize to complex real-world indoor scenes.} 
    \label{fig:understanding_all_results}
\end{figure*}

\begin{table*}[ht]
    \centering
    \scriptsize
    \renewcommand{\arraystretch}{0.75}
    \resizebox{\textwidth}{!}{%
    \begin{tabular}{l|l|c|cccccccc}
        \toprule
        \textbf{Supervision} & \textbf{Method} & \textbf{Input}
        & \textbf{EM@1} $\uparrow$ 
        & \textbf{B-1} $\uparrow$
        & \textbf{B-2} $\uparrow$ 
        & \textbf{B-3} $\uparrow$ 
        & \textbf{B-4} $\uparrow$ 
        & \textbf{METEOR} $\uparrow$ 
        & \textbf{ROUGE-L} $\uparrow$ 
        & \textbf{CIDEr} $\uparrow$ \\
        \midrule
        \multirow{12}{*}{\textbf{Training}} 
        & ScanQA     & \multirow{12}{*}{\centering PC} & 21.05 & 30.24 & 20.40 & 15.11 & 10.08 & 13.14 & 33.33 & 64.86 \\
        & 3DMIT      &  & 13.04     & 27.63     & -     & -     & 5.24 & 10.70     & 26.22     & 48.03 \\
        & 3D-LLM(Flamingo)      &  & 20.40 & 30.30 & 17.80 & 12.00 & 7.20 & 12.20 & 32.30 & 59.20 \\
        & 3D-LLM(BLIP2-flant5)  &  & 20.50 & 39.30 & 25.20 & 18.40 & 12.00 & 14.50 & 35.70 & 69.40 \\
        & 3D-VisTA             &  & 22.40     & - & - & -     & 10.40     & 13.90 & 35.70 & 69.60 \\
        & LL3DA                &  & -     & -     & -     & -     & 13.53 & 15.88 & 37.31 & 76.79 \\
        & Scene-LLM            &  & 27.20 & \textbf{43.60} & 26.80 & 19.10 & 12.00 & 16.60 & 40.00 & 80.00 \\
        & Chat-Scene           &  & 21.62 & 43.20 & \textbf{29.06} & 20.57 & 14.31 & \textbf{18.00} & \textbf{41.56} & 87.70 \\
        & 3D-VLP               &  & 21.65 & 30.53 & 21.33 & 16.67 & 11.15 & 13.53 & 34.51 & 66.97 \\
        & Grounded 3D-LLM      &  & -     & -     & -     & -     & 13.40 & -     & -     & 72.70 \\
        & Chat-3D              &  & -     & 29.10 & -     & -     & 6.40 & 11.90 & 28.50 & 53.20 \\
        & Chat-3D v2           &  & 21.10 & 38.40     & -     & -     & 7.30 & 16.10 & 40.10 & 77.10 \\
        \midrule
        \rowcolor[gray]{0.95}
        \textbf{Training Free} & \textbf{Sparse3DPR(Ours)} & SVI
        & \textbf{27.22} & 36.23 & 28.42 & \textbf{22.92} & \textbf{14.99} & 17.40 & 37.98 & \textbf{88.07} \\
        \bottomrule
    \end{tabular}
    }
    \caption{\textbf{Evaluations on ScanQA.} Our training-free method, operating on sparse-view images (SVI), is benchmarked against training-based approaches that use point clouds (PC). Best results are highlighted in bold. B-1 to B-4 are BLEU-n scores. Sparse3DPR achieves state-of-the-art results on several metrics (EM@1, BLEU-3, BLEU-4, and CIDEr).}
    \label{tab:scanqa_benchmark_comparison}
\end{table*}

\subsubsection{3D Scene Question Answering.} In this section, we evaluate Sparse3DPR on the ScanQA validation set \cite{azuma2022scanqa}, compare it with state-of-the-art (SOTA) approaches \cite{azuma2022scanqa, li20243dmit, hong20233d, chen2024ll3da, fu2024scene-llm, huang2024chatscene, jin20233D-VLP, chen2024grounded-3d-llm, wang2023chat-3d} that rely on 3D-specific inputs like point clouds. As shown in Table~\ref{tab:scanqa_benchmark_comparison}, Sparse3DPR demonstrates highly competitive performance. It achieves SOTA results on several key metrics, including an EM@1 of 27.22\% and a CIDEr score of 88.07\%, which respectively reflect its factual accuracy and ability to generate semantically rich responses. Notably, this strong performance is achieved in a training-free, zero-shot setting using only sparse RGB inputs. This success is attributed to our framework's ability to provide the LLM with a superior context, where our HPSG establishes a robust and spatially coherent structural foundation, and our task-adaptive subgraph extraction ensures this context is query-relevant and redundancy-free for more accurate and reliable reasoning.

\begin{table*}[ht]
    \centering
    \large
    \renewcommand{\arraystretch}{1.0}
    \resizebox{\textwidth}{!}{%
    \begin{tabular}{l|c|c|cccccccc|c}
        \toprule
        \multirow{2}{*}{\textbf{Method}} 
        & \multirow{2}{*}{\textbf{Subgraph}}  
        & \multirow{2}{*}{\textbf{SG Type}}  
        & \multirow{2}{*}{\textbf{EM@1} $\uparrow$} & \multirow{2}{*}{\textbf{B-1} $\uparrow$} & \multirow{2}{*}{\textbf{B-2} $\uparrow$} & \multirow{2}{*}{\textbf{B-3} $\uparrow$} & \multirow{2}{*}{\textbf{B-4} $\uparrow$} & \multirow{2}{*}{\textbf{METEOR} $\uparrow$} & \multirow{2}{*}{\textbf{ROUGE-L} $\uparrow$} & \multirow{2}{*}{\textbf{CIDEr} $\uparrow$}
        & \textbf{Inference} \\
        & & & & & & & & & & & \textbf{Time (avg)} $\downarrow$ \\
        \midrule
        ConceptGraphs     & \textcolor{red}{\ding{55}} & Flat     & 20.23 & 11.72 & 6.64 & 3.79 & 2.18 & 11.05 & 35.54 & 85.43  & 2.15s \\
        ConceptGraphs*    & \textcolor{red}{\ding{55}} & Flat     & 26.94 & 20.92 & 12.11 & 7.38 & 4.47 & 15.82 & 49.69 & 115.79 & 1.47s \\
        \rowcolor[gray]{0.95}
        Sparse3DPR*            & \textcolor{red}{\ding{55}} & Flat     & 28.97 & 21.42 & 12.22 & 7.55 & 4.55 & 15.96 & 49.90 & 133.38 & 1.08s \\
        \rowcolor[gray]{0.95}
        Sparse3DPR$^{\dagger}$ & \textcolor{red}{\ding{55}} & Afford.  & 27.17 & 21.51 & 12.69 & 8.02 & 4.99 & 15.45 & 47.67 & 122.30 & 1.02s \\
        \rowcolor[gray]{0.95}
        Sparse3DPR*            & \textcolor{red}{\ding{55}} & HPSG     & \textbf{30.91} & \textbf{24.03} & \textbf{13.98} & \textbf{8.86} & \textbf{5.64} & \textbf{16.51} & \textbf{51.81} & \textbf{141.15} & \textbf{0.94s} \\
        \midrule
        ConceptGraphs     & \textcolor{green}{\ding{51}} & Flat     & 28.04 & 27.71 & 17.14 & 10.94 & 6.18 & 14.89 & 49.06 & 126.00 & 0.39s \\
        ConceptGraphs*    & \textcolor{green}{\ding{51}} & Flat     & 32.49 & 27.14 & 16.53 & 10.61 & 6.46 & 15.50 & 52.22 & 140.49 & 0.39s \\
        \rowcolor[gray]{0.95}
        Sparse3DPR*            & \textcolor{green}{\ding{51}} & Flat     & 32.82 & 30.43 & 19.08 & 12.85 & 8.16 & 16.55 & 52.28 & 146.39 & 0.38s \\
        \rowcolor[gray]{0.95}
        Sparse3DPR$^{\dagger}$ & \textcolor{green}{\ding{51}} & Afford.  & 30.78 & 29.00 & 17.54 & 11.41 & 7.08 & 15.13 & 50.98 & 136.08 & 0.37s \\
        \rowcolor[gray]{0.95}
        \textbf{Sparse3DPR (Ours)} & \textcolor{green}{\ding{51}} & HPSG  & \textbf{34.68} & \textbf{31.13} & \textbf{19.90} & \textbf{13.60} & \textbf{8.87} & \textbf{17.56} & \textbf{53.17} & \textbf{160.73} & \textbf{0.32s} \\
        \bottomrule
    \end{tabular}
    }
    \caption{\textbf{Ablation study on Space3D-Bench.} All compared methods are training-free. “ConceptGraphs” and “ConceptGraphs*” serve as flat SG baselines constructed following prior methods. Sparse3DPR* denotes variants of Sparse3DPR either without the subgraph extraction method or with flat SGs obtained by flattening HPSG. Sparse3DPR$^{\dagger}$ reorganizes HPSG into an affordance-based hierarchical structure. The \textit{SG Type} column specifies the organization of the scene graph: Flat (non-hierarchical), Afford. (affordance-based hierarchy), or HPSG (hierarchical plane-enhanced scene graph).}
    \label{tab:3D_bench_qa_results}
\end{table*}

\subsubsection{Qualitative Analysis of 3D Scene Understanding.}
To demonstrate the practical applicability and generalization of our method, we conduct qualitative experiments in a real-world laboratory scene characterized by cluttered object distributions and irregular layouts. This environment presents significant challenges for prior paradigms that are often constrained by domain gaps or a reliance on 3D-specific data such as point clouds or dense posed RGB-D sequences. As demonstrated in Figure~\ref{fig:understanding_all_results}, Sparse3DPR effectively handles complex queries spanning object understanding, spatial reasoning, and geometric analysis under these conditions. This robust performance stems from the framework's ability to efficiently parse the scene, extract task-relevant context such as captions, 3D positions, and geometric properties, and integrate this information into structured prompts for context-aware reasoning. Crucially, the entire process operates without task-specific fine-tuning and relies solely on sparse-view RGB inputs, demonstrating strong generalization and practical deployability. For more extensive qualitative examples, please refer to the supplementary material.

\subsection{Ablation Study}
To evaluate the impact and effectiveness of HPSG and our task-adaptive subgraph extraction, we conduct ablation studies on 3D QA using a seven-scene subset of Space3D-Bench. For this analysis, we build flat SG baselines using the SG from ConceptGraphs, and further reorganize HPSG into two alternative designs for comparison: a flat structure consistent with ConceptGraphs and an affordance-based hierarchical structure following TB-HSU \cite{xu2025tbHSU}. To ensure a fair comparison, all variants are evaluated using the identical reasoning pipeline, including the same LLM and prompt template. Additional details on baselines and experimental settings are available in the supplementary material. The results are summarized in Table \ref{tab:3D_bench_qa_results}.
 
\subsubsection{Impact of SG type.} To isolate the impact of different SGs on reasoning performance, we disable the task-adaptive subgraph extraction and feed the full graph to the LLM. As shown in the top block of Table~\ref{tab:3D_bench_qa_results}, Sparse3DPR* (Flat SG) outperforms the baseline ConceptGraphs* (Flat SG) with higher EM@1, indicating the high quality of our underlying scene parsing. In contrast, Sparse3DPR$^{\dagger}$ (Afford. SG) employs a function-centric hierarchy that shifts the LLM’s focus from explicit objects and spatial relations to object affordances, improving fluency and reasoning efficiency at the cost of accuracy (EM@1). Our Sparse3DPR* (HPSG), however, uses a physically-grounded hierarchy that preserves spatial coherence, leading to significant improvements in both EM@1 accuracy and reasoning speed. This proves that a human-intuitive, spatially coherent SG structure is more effective for LLM-based reasoning.

\subsubsection{Impact of task-adaptive subgraph extraction.} We first analyze the primary impact of our task-adaptive subgraph extraction. As shown in Table \ref{tab:3D_bench_qa_results} (bottom vs. top blocks), enabling this method yields a significant and consistent performance improvement across all SG types in both reasoning accuracy and efficiency. These results demonstrate that dynamically providing the LLM with a focused, query-relevant, and noise-free context is critical for effective reasoning. Furthermore, these results also reveal an interplay between the subgraph extraction method and the underlying SG structure (as shown in the bottom block). The performance benefit is most pronounced when combined with our HPSG, which achieves the best results across all metrics (e.g., 0.32s reasoning time). In contrast, when the subgraph is combined with the Affordance-based SG, performance declines. We attribute this to its function-centric hierarchy, which emphasizes functional relationships, leading to inaccurate seed node selection and the inclusion of irrelevant scene context. By comparison, HPSG preserves spatial coherence through dominant planar structures as spatial anchors, enabling precise seed node retrieval and producing cleaner, task-relevant subgraphs for improved performance.

\section{Conclusion}
We propose Sparse3DPR, a novel framework that addresses the challenges of accuracy and efficiency in the practical deployment of training-free 3D scene understanding. By constructing an HPSG from sparse-view RGB inputs, it provides a spatially coherent, semantically rich, and reasoning-friendly representation, while the task-adaptive subgraph extraction method dynamically filters redundant context and retains task-relevant information, thereby improving reasoning accuracy and efficiency. Experimental results show that Sparse3DPR significantly improves both accuracy and speed over previous training-free methods and achieves comparable performance to training-based counterparts on the ScanQA benchmark, further confirming its efficiency and generalizability for real-world applications. Future work will focus on extending this framework to temporal reasoning in dynamic environments.


\bibliography{aaai2026}

\clearpage


\section*{Supplementary material}

# Sparse3DPR: Training-Free 3D Hierarchical Scene Parsing and Task-Adaptive Subgraph Reasoning from Sparse RGB Views

# Supplementary Materials

## 1 Additional Experimental Details

### 1.1 Datasets and Evaluation Metrics

*Replica*: A dataset of high-fidelity 3D indoor environments with dense geometry and semantic labels (Straub et al. 2019). We use a subset of eight scenes, comprising five offices (offices 0,1,2,3,4) and three rooms (rooms 0,1,2), to evaluate the parsing accuracy of the scene entities generated by our Sparse3DPR framework. These entities form the foundational nodes of the resulting hierarchical plane-enhanced scene graph (HPSG). This accuracy is quantitatively evaluated using two standard 3D semantic segmentation metrics: mAcc and F-mIoU.

*ScanQA*: A large-scale 3D visual question answering (3D-VQA) benchmark grounded in real-world indoor scans, featuring a diverse set of free-form questions that require spatial and semantic reasoning (Azuma et al. 2022). Our evaluation is performed on the official validation split to assess the model's capability on this reasoning task. Specifically, we evaluate the generated answers using several standard metrics, including exact match at top-1 (EM@1), BLEU-n (B-1 to B-4), ROUGE-L, METEOR, and CIDEr.

*Space3D-Bench*: A high-quality, diverse 3D spatial question answering dataset (Szymańska et al. 2025), derived from the Replica dataset and designed for 3D spatial reasoning and understanding. We utilize the QA pairs from a curated subset of seven scenes, encompassing four offices (offices 0,2,3,4) and three rooms (rooms 0,1,2). Furthermore, guided by the ScanQA question typology, we refine these QA pairs by removing all questions requiring precise quantitative measurements, thereby focusing the evaluation on qualitative spatial and semantic reasoning. The evaluation is conducted using the same set of metrics as used for ScanQA.

### 1.2 Experimental Settings

In this work, we use Qwen-VL (Qwen2.5-VL-7B-Instruct) (Bai et al. 2025) for image captioning and language reasoning, and DUSt3R (DUSt3R-512-DPT) (Wang et al. 2024) for sparse-view 3D geometry extraction. For image segmentation, we adopt SAM2 (sam2.1_hiera_large) (Ravi et al. 2024), while open-vocabulary object detection is handled by RAM++ (ram_plus_swin_large_14m) (Huang et al.

2023) and GroundingDINO (groundingdino_swint_ogc) (Liu et al. 2024). To obtain text embeddings, we employ SentenceTransformer (all-mpnet-base-v2) (Reimers and Gurevych 2019). For data sampling, we uniformly sample 28 frames per scene from Replica-based datasets, including Replica (Straub et al. 2019) and Space3D-Bench (Szymańska et al. 2025), and 22 frames per scene from ScanQA (Azuma et al. 2022). All experiments are conducted on a single workstation equipped with an NVIDIA GeForce RTX 4090 GPU.

### 1.3 Ablation Baseline Implementation

This section details the implementation of the baseline methods and the specific variants of our HPSG for comparison.

**Baseline methods.** We establish two foundational baselines derived from the scene graph (SG) generation methods in ConceptGraphs (Gu et al. 2024): one that uses class-agnostic instance masks (from SAM) to generate nodes, and another using a dedicated open-vocabulary object detector. Since the original ConceptGraphs framework is not designed for the 3D question answering (QA) task, we integrate its generated SGs from both methods into our reasoning pipeline. To ensure fair and robust comparison, each baseline's SG is evaluated under two configurations: (1) directly using the full SG, and (2) applying our task-adaptive subgraph extraction prior to reasoning. Across all configurations, the reasoning components, including prompt templates and the LLM, remain identical to those used in our proposed Sparse3DPR framework. This setup ensures a controlled comparison where only the input SG structure varies.

**HPSG Structural Variants.** To analyze the impact of our novel hierarchical design, we create two distinct structural variants of the HPSG for an extensive ablation study. For a fair comparison, all other aspects, including node attributes and edge information, are kept identical.
*HPSG (Flat)*: This variant flattens the multi-level HPSG into a single-layer SG. We achieve this by removing the top-level scene-type node and merging all structural and object nodes into one level, while preserving their original topological and spatial relationships. This resulting flattened structure is analogous to prior work like ConceptGraphs.
*HPSG (Affordance-based):* Following the design principles of TB-HSU (Xu et al. 2025), this variant restructures the

graph's second level from a structure-centric into a function-centric hierarchy. Specifically, we use an LLM to interpret the captions of structural nodes and generate abstract functional concepts (e.g., "for resting"). These new functional nodes are inserted at the graph's second level. The original structural nodes are then moved to the third level and re-linked as children of their corresponding functional parent, effectively creating an affordance-based hierarchy.

### 1.4 Prompt Template Design

To ensure consistent and controllable outputs from the LLM for reasoning tasks, we employ a set of structured prompt templates. The specific design for each is detailed below:

*Spatial relationship reasoning prompt*: To infer the spatial relationship between two objects, we utilize an LLM guided by a prompt that leverages key object information to generate a spatial relationship description. The prompt, illustrated in Figure 3, constrains the LLM to output only a single keyword that best represents the spatial relationship, selected from a predefined set (e.g., on, in, next_to).

*Detailed 3D-QA prompt for ScanQA dataset*: To ensure consistent and high-quality LLM responses for the 3D QA task on ScanQA, we design a structured few-shot prompt, as illustrated in Figure 4. The prompt is based on a systematic analysis of the ScanQA training set, capturing recurring patterns in question phrasing and answer style. Our method is evaluated exclusively on the validation set.

*Detailed 3D-QA prompt for Space3D-Bench dataset*: The few-shot prompt designed for the Space3D-Bench dataset is illustrated in Figure 5. Its exemplars are carefully curated from scenes not included in the test set, ensuring that the prompt remains generalizable. For consistency and a fair comparison in our ablation studies, this exact prompt is used across all evaluated methods, including our approach and all baselines. This ensures that the performance evaluation is conducted under the same conditions, allowing for a meaningful comparison of results.

## 2 Extended Ablation Studies

To analyze the impact of view sparsity, we conducted a sensitivity analysis on Space3D-Bench by varying the number of input views from 4 to 28. As shown in Table 1, the results indicate a clear trend where performance generally improves with an increasing number of views. We observed that performance with 16 views slightly surpasses that with 22 views, which we attribute to the higher informational quality and reduced redundancy of the 16-view set. This configuration likely captures the scene's essential geometric and semantic information with optimal efficiency. In contrast, the 22-view set may introduce redundant or lower-quality scene information, leading to a marginally less accurate 3D scene representation. This finding underscores that, beyond a certain threshold, the quality and coverage of views are more critical than their quantity. Meanwhile, the best performance at 28 views indicates that increasing the number of sparse views is beneficial, provided these views offer new and valuable information. Therefore, we conclude that for comprehensive scene understanding, dense visual input is not strictly necessary. Instead, a sparse yet well-distributed set of views that effectively captures the scene's key geometric and semantic features is more advantageous.

## 3 Additional Qualitative Results

To further demonstrate the generalization and robustness of Sparse3DPR, we present additional qualitative results across diverse environments, including high-fidelity synthetic scenes from the Replica dataset (Figure 1) and a challenging real-world laboratory scan (Figure 2). As illustrated, Sparse3DPR effectively interprets a wide range of user instructions, including descriptive queries for object understanding (a), complex questions about relative distances and positions for spatial reasoning (b), and size-based analysis for geometric understanding (c). The framework's ability to generate accurate, detailed responses across these varied scenes and query types, all from sparse RGB inputs, demonstrates its strong generalization and real-world applicability.

| Method | Views | EM@1 ↑ | B-1 ↑ | B-2 ↑ | B-3 ↑ | B-4 ↑ | METEOR ↑ | ROUGE-L ↑ | CIDEr ↑ |
|---|---|---|---|---|---|---|---|---|---|
| Sparse3DPR (Ours) | 28 | **34.68** | **31.13** | **19.90** | **13.60** | **8.87** | **17.56** | **53.17** | **160.73** |
| | 22 | 31.58 | 28.18 | 17.80 | 11.76 | 7.00 | 15.62 | 51.65 | 140.40 |
| | 16 | 31.91 | 29.20 | 18.77 | 12.56 | 7.38 | 16.07 | 51.55 | 141.18 |
| | 10 | 30.62 | 28.57 | 18.26 | 12.20 | 7.56 | 15.31 | 49.54 | 143.15 |
| | 4 | 27.85 | 26.25 | 16.64 | 11.21 | 6.76 | 13.56 | 47.17 | 128.70 |

Table 1: **Sensitivity analysis of Sparse3DPR to the number of input views on Space3D-Bench.** Performance generally improves with an increasing number of views, with the best results achieved at 28 views. Best results are in bold.

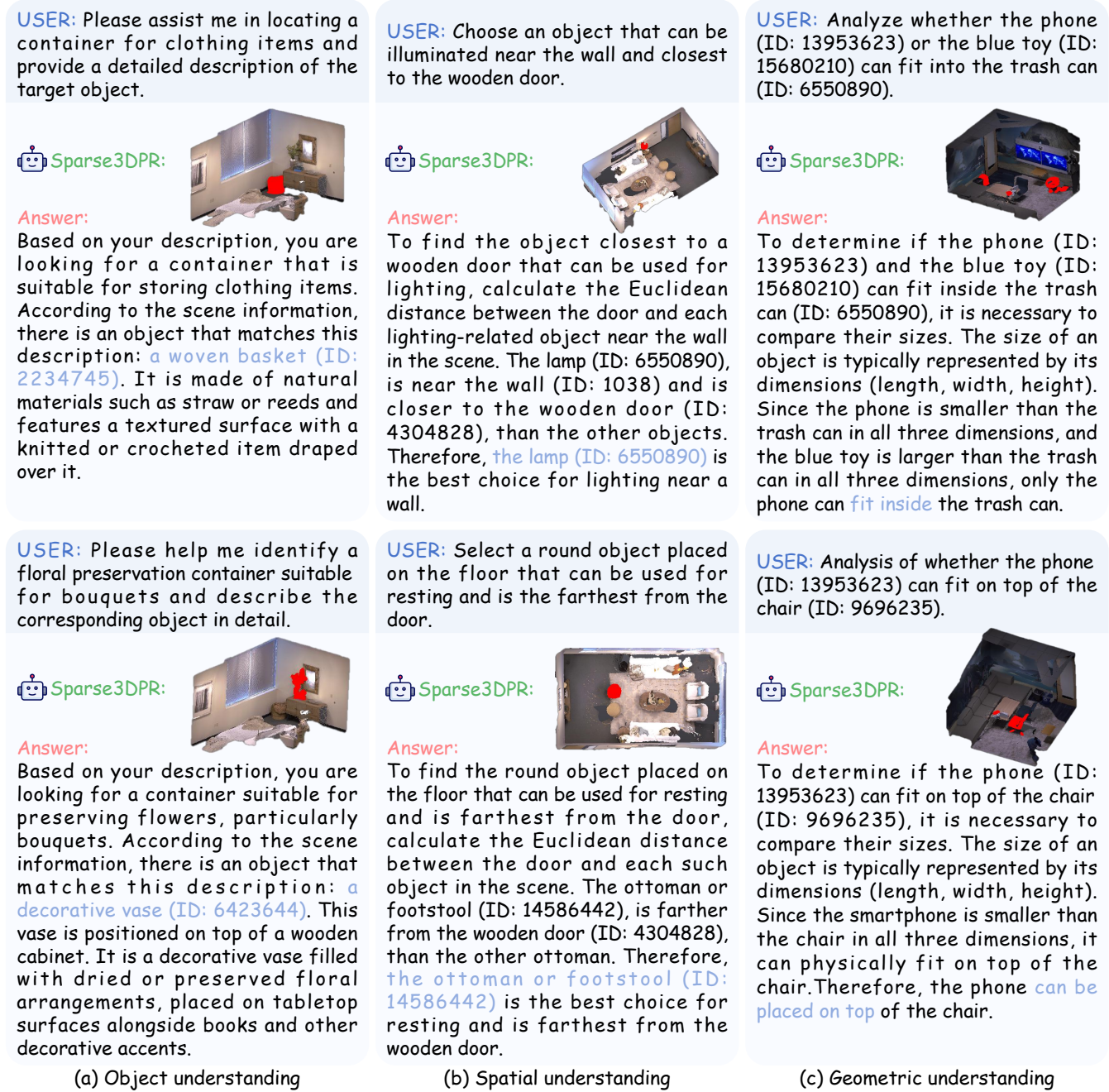

(a) Object understanding  (b) Spatial understanding  (c) Geometric understanding

Figure 1: **Qualitative results of Sparse3DPR on Replica scenes**

| USER: Please look for a computer that is easy to carry, works well in various environments. | USER: Choose a computer monitor near the wall that is closest away from the wooden door. | USER: Analysis of whether the computer monitor (ID: 2630964) can fit inside the AOC box (ID: 4998151). |
|---|---|---|
| 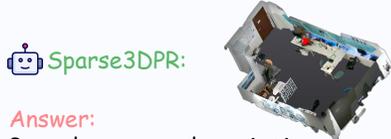 Sparse3DPR: | 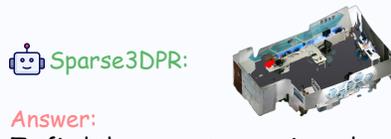 Sparse3DPR: | 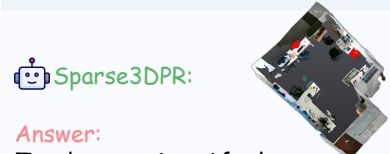 Sparse3DPR: |
| Answer: Based on your description, you are looking for a computer that is easy to carry and works well in various environments. According to the scene information, there is an object that matches this description: a laptop computer (ID: 13072738). This laptop is located on a table, positioned at [-0.471824, -0.097297, -0.031946]. It is placed on a desk and is accompanied by a black bag or pouch. | Answer: To find the computer monitor closest to the wooden door, calculate the Euclidean distance between the door and each monitor near the wall in the scene. The computer monitor (ID: 3747645), located at [-0.519387, -0.055954, -0.041962], is closer to the wooden door (ID: 11525593), located at [-0.653927, 0.149246, -0.052542], than the other monitor. Therefore, the computer monitor (ID: 3747645) is the closest to the wooden door. | Answer: To determine if the computer monitor (ID: 2630964) can fit inside the AOC box (ID: 4998151), it is necessary to compare their sizes. The size of an object is typically represented by its dimensions (length, width, height). Since the height of the computer monitor (0.038603) is greater than the height of the AOC box (0.019202). Therefore, the computer monitor does not fit inside the AOC box. |
| (a) Object understanding | (b) Spatial understanding | (c) Geometric understanding |

Figure 2: **Qualitative results of Sparse3DPR on a real-world laboratory scene.**

---

Spatial Relationship Reasoning Prompt

Analyze the spatial relationship between two objects. Given the 3D bounding box centers and dimensions of two objects, determine the most likely spatial relationship between them.

Please return one of the following relationships:

- `"on"`: if object 1 is typically placed on top of object 2
- `"supports"`: if object 1 typically supports object 2 (object 2 is on top of object 1)
- `"in"`: if object 1 is typically inside object 2
- `"contains"`: if object 1 typically contains object 2 (object 2 is inside object 1)
- `"next_to"`: if object 1 and object 2 are typically placed side by side
- `"none"`: if none of the above relationships apply

Return only a single word as the answer, with no additional text.

Figure 3: **The prompt example for spatial relationship reasoning.**

**3D Scene Analysis Assistant Prompt**

**System Role:** You are a 3D scene analysis assistant. Your task is to answer questions about objects in a room based on the provided scene information.

**CRITICAL INSTRUCTIONS FOR YOUR RESPONSE (Follow these VERY STRICTLY):**

1. **Extreme conciseness & target length:** Your answer MUST be as short as possible, typically 1–5 words (e.g., *"dark brown"*, *"on wall"*, *"2"*, *"tv"*).
2. **Directness & focus on retrieved info:** Answer the question directly based *only* on the `Retrieved scene information`.
3. **No prefixes/suffixes/IDs/punctuation:** Do NOT use prefixes like *"The object is:"*. Do NOT end with a period. Never include raw object IDs.
4. **Match question type and examples precisely:**
   - Use simplest lowercase names (e.g., *"chair"*, *"table"*, *"tv"*).
   - If color/attribute is essential, include it concisely (e.g., *"black tv"*).
   - For `How many` questions, respond with a number + object type (*"2 silver monitors"*) or just a number (*"2"*).
   - Prioritize conciseness and consistency with example style.
5. **Object descriptions (when concise & relevant):** Use the most concise and relevant description from the context. Prioritize conciseness.
6. **No explanations or filler:** Provide ONLY the answer. No reasoning, apologies, or extra text.

**FORMAT EXAMPLES BASED ON QUESTION TYPE (answers must be lowercase):**

**If question starts with *"Where"*:**
(Example answers: on, under, in front of, left, to left, to right, above, next to, on right side, on top of, right)
**Q:** Where is a small end table in the corner? **A:** right of couch

**If question starts with *"How many"*:**
(Example answers: zero, one, two, three, four, five, six, seven, eight, nine, ten)
**Q:** How many tables is the couch sandwiched between? **A:** 2 tables

**If question starts with *"What color"* or *"What is the color"*:**
(Example answers: white, brown, black, blue, grey, red, tan, light brown, gray, beige, green, dark brown, silver, yellow, black chair, orange, dark gray, metallic gray)
**Q:** What color is the cover over the piano? **A:** red

**If question starts with *"What shape"*, *"What type"*, *"What kind"*, or *"What ... made of"*:**
(Example answers: rectangular, rectangle, square, rectangular shape, round, wooden, oval, whiteboard, black, circular, cylindrical, glass, wood, metal, long gray counter, brown plastic, rectangular and brown, padded gray, metal base)
**Q:** What type of table is in the middle of the room? **A:** dark colored dining room table

**If question starts with *"What is"* (object identification):**
(Example answers: chair, table, trash can, window, desk, door, couch, picture, whiteboard, cabinet, shelf, radiator, bed, coffee table, lamp, sink, chairs, tv, fuse box, pipe, towel, furnace)
**Q:** What is one of several at a table? **A:** chair

**Yes/No questions: Q:** Does the black office chair by the desk have wheels? **A:** yes **(Respond with ONLY 'yes' or 'no', in lowercase.)**

**For other question types (e.g., starting with "Which", or not fitting above categories):**
(Example answers: right, kitchen utensils, radiator, cabinet, left, chair, table, window, desk, shelf, wall, couch, door)
Provide the most direct and concise lowercase answer possible, similar in style to all examples above.

**Retrieved scene object information:** {scene_context}
**User question:** {input_query}
**Answer (follow ALL instructions and examples, be extremely concise, lowercase, no IDs):**

Figure 4: **Detailed prompt for ScanQA.** It employs a comprehensive set of instructions and dynamic placeholders ({*scene_context*}, {*input_query*}) to control the style and content of the LLM's responses, ensuring conciseness and adherence to the provided context.

> **3D Scene Analysis Assistant Prompt**
>
> You are a 3D scene analysis assistant answering questions about objects in a room.
> **CRITICAL FORMATTING INSTRUCTIONS:** Your answer format **must** match the exact example formats shown below. Answers must be extremely concise.
> **Formatting guide based on question type:**
>
> - **Yes/No questions**
>   If the question starts with *"Are there"*, *"Is there"*, or *"Could you"*: Respond with only **"Yes"** or **"No"**. No explanation.
> - **Counting questions**
>   If the question contains *"How many"*:
>   - If referring to people: respond with **"Number of people: X"**
>   - Otherwise: respond with **"Number of objects: X"**
>
>   (Replace X with the appropriate digit. No explanation.)
> - **Object identification**
>   If the question starts with *"What is closest"*: Respond with **"Object: X"**, where X is the object name. No explanation.
> - **Descriptive questions**
>   For all other descriptive questions, provide a concise yet thorough answer (2–4 sentences) that:
>   1. Includes detailed information about all relevant objects
>   2. Describes physical attributes (color, size, shape, material)
>   3. Mentions spatial relationships between objects
>   4. Covers all aspects mentioned in the retrieved information
>
> **FORMAT EXAMPLES BY QUESTION TYPE:**
> **Yes/No Questions:**
>   **Q:** "Are there any windows?" **A:** No
> **Counting Questions:**
>   **Q:** "How many chairs are there in the apartment?" **A:** Number of objects: 6
> **Object Identification:**
>   **Q:** "What is closest to the white sofa?" **A:** Object: table
> **Descriptive Questions:**
>   **Q:** "Describe the objects in the kitchen" **A:** The kitchen includes a double-door fridge, white drawers and cabinets, a countertop with a sink, a dishwasher, and an oven.
> **Retrieved scene object information:** {scene_context}
> **User question:** {input_query}
> **Answer (MAINTAIN EXACT FORMAT from examples):**

Figure 5: **Few-shot prompt designed for Space3D-Bench.** It standardizes LLM outputs for automated evaluation by defining strict formatting rules and using placeholders ({*scene_context*} and {*input_query*}) that are dynamically populated at runtime.


\end{document}